\documentclass[letter,12pt]{article}
\usepackage{newtxtext,newtxmath, amsmath, amsfonts, bm,mathtools}
\usepackage{paralist, url, natbib, parskip}
\usepackage{adjustbox, todonotes} 
\usepackage{graphicx, subfig, rotating, booktabs, multirow}
\usepackage[makeroom]{cancel}
\usepackage[section]{placeins}
\usepackage[linesnumbered,lined,boxed,commentsnumbered]{algorithm2e}
\usepackage{xcolor, tikz}
\usetikzlibrary{decorations.markings}
\usetikzlibrary{patterns}
\usetikzlibrary{shapes.geometric, arrows}
\usetikzlibrary{calc}
\usetikzlibrary{arrows.meta}
\usetikzlibrary{positioning}
\usetikzlibrary{shapes}
\usetikzlibrary{fit}
\usepackage[margin=3cm]{geometry}
\usepackage[margin=2cm]{caption}
\usepackage{titling}
\usepackage{setspace}
\DeclareCaptionStyle{italic}[justification=centering]{labelfont={bf},textfont={it},labelsep=colon}
\SetKwInOut{Parameter}{parameter}
\DeclareMathOperator{\sign}{sign}
\mathtoolsset{showonlyrefs}

\newcommand{\cor}{\mathop{\text{cor}}}

\graphicspath{{./Graphics/}}

\tikzstyle{person} = [rectangle, rounded corners, minimum width=2cm, minimum height=0.5cm,text centered, draw=black, fill=blue!30]
\tikzstyle{testitem} = [trapezium, trapezium left angle=80, trapezium right angle=110, minimum width=1cm, minimum height=0.5cm, text centered, draw=black, fill=green!30]
\tikzstyle{model} = [rectangle, minimum width=2cm, minimum height=1cm, text centered, draw=black, rounded corners, fill=orange!30]
\tikzstyle{object} = [rectangle, minimum width=2cm, minimum height=2cm, text centered, draw=black,  text width=2cm, fill=brown!30]
\tikzstyle{decision} = [diamond, minimum width=3cm, minimum height=1cm, text centered, draw=black, fill=green!30]
\tikzstyle{arrow} = [thick,->,>=stealth]
\tikzstyle{line} = [thick,-,>=stealth]
\tikzstyle{std} = [rectangle, rounded corners, minimum width=2cm, minimum height=1cm,text centered,text width= 3cm, draw=black]

\usepackage{array, ragged2e}
\newcolumntype{P}[1]{>{\raggedright\arraybackslash}p{#1}}
\RaggedRight
\usepackage{adjustbox}
\newcolumntype{R}[2]{%
    >{\adjustbox{angle=#1,lap=\width-(#2)}\bgroup}%
    l%
    <{\egroup}%
}

\allowdisplaybreaks
\begin{document}
\tikzset{->-/.style={decoration={
  markings,
  mark=at position #1 with {\arrow{>}}},postaction={decorate}}}

\title{Unsupervised Anomaly Detection Ensembles using Item Response Theory}
\author{Sevvandi Kandanaarachchi$^1$}
\date{%
   \scriptsize{ $^1$ School of Science,  Mathematical Sciences, RMIT University, Melbourne VIC 3000, Australia.\\ 
   [2ex]}%
    }
\begin{titlingpage}
\maketitle
\begin{abstract}
Constructing an ensemble from a heterogeneous set of unsupervised anomaly detection methods is challenging because the class labels or the ground truth is unknown. Thus, traditional ensemble techniques that use the response variable or the class labels cannot be used to construct an ensemble for unsupervised anomaly detection. 

We use Item Response Theory (IRT) -- a class of models  used in educational psychometrics to assess student and test question characteristics -- to construct an unsupervised anomaly detection ensemble. IRT's latent trait computation lends itself to anomaly detection because the latent trait can be used to uncover the hidden ground truth. Using a novel IRT mapping to the anomaly detection problem, we construct an ensemble that can downplay noisy, non-discriminatory methods and accentuate sharper methods. We demonstrate the effectiveness of the IRT ensemble on an extensive data repository, by comparing its performance to other ensemble techniques. 
\end{abstract}


\end{titlingpage}

\newgeometry{top=1.5cm,bottom=2cm,right=1.5cm,left=1.5cm}
\section{Introduction}\label{sec:intro}
Unsupervised Anomaly Detection (AD) is used in diverse societal applications such as identifying fraudulent credit card transactions and  intrusions in computer networks making it critical to correctly identify anomalies minimizing  costly false positives and dangerous false negatives. Ensembles have proven effective in increasing the performance accuracy of  many algorithm classes.
A significant challenge in unsupervised anomaly detection is that we do not know the anomalies, that is, the ground truth or the class labels are unknown. Because of this, the proven and tested ensemble frameworks that use the class labels cannot be applied for anomaly detection. We propose a novel ensemble framework based on Item Response Theory (IRT), which models the ground truth as a latent trait. 

Unsupervised anomaly detection ensembles can be broadly categorized into 3 types based on their ensemble strategy. These are subsampling, feature bagging and the use of different combination functions \citep{Aggarwal2015}. Subsampling strategies \citep{Zimek2013} generally use different samples of observations from the dataset to build a model and average the predictions from various observation samples. Feature bagging \citep{Lazarevic2005} uses different subsets of features of the dataset to build the model. Again the prediction scores are averaged. Generally, both subsampling and feature bagging use the same base algorithm on different subsets of observations and/or features to produce a final consensus. The use of different combination functions such as the average, maximum and other correlation based functions have also been explored in ensemble learning. Combination function ensemble methods are generally used with a set of heterogeneous anomaly detectors.  The proposed IRT ensemble framework in effect constitutes a new combination function that can be used on set of heterogeneous AD methods. 


Combination functions are often used in supervised machine learning ensembles. For example, gradient boost algorithms combine the output of a series of models, each predicting the residuals of a prior model.  It is difficult to emulate such a process in the unsupervised AD ensemble regime, because the class labels are not known. Some AD ensemble techniques use  pseudo-ground-truth labels \citep{Rayana2016, Campos2018, Zhang2019}, which are generally defined as averaged anomaly scores, in place of the unknown ground truth.  
The use of pseudo-ground-truth labels in AD ensembles gives rise to a circular argument because 
the ensemble  mimics the pseudo-ground-truth labels, which affects the ensemble nuts and bolts in terms of the choice of AD methods and their weights. Therefore, if the pseudo-ground-truth labels are not true representatives of the actual ground-truth, which often they are not, their use constitutes a serious limitation.


The proposed ensemble method steers away from the limitations of pseudo-ground-truth by modeling the ground truth as a latent variable. Latent variables are variables which are not directly observed, but inferred from a mathematical model. Thus, they are hidden. For example, the market volatility can be modeled as a latent variable using the observed stock prices; a student's ability can be modeled as a latent variable using the student's responses to a test. Similarly, the unknown ground truth of anomalies can be modeled as a latent variable using the observed scores of AD methods. 

Item Response Theory (IRT) encompasses a family of latent trait models that has traditionally been used in educational psychometrics to determine the ability of test participants and the difficulty and discrimination of test questions.   Recently, IRT has been used for algorithm and dataset evaluation \citep{Martinez-Plumed2019, Chen2020, skksm2020}, and  classification ensemble construction \citep{Chen2020ensemble} broadening the applicability of IRT to the machine learning and artificial intelligence domain. Of particular interest is the IRT ensemble proposed by \cite{Chen2020ensemble}. Their ensemble, which was used for classification algorithms can be also extended for other supervised machine learning algorithms. However, it cannot be extended to unsupervised tasks because they use accuracy measures in their IRT model. That is, they assign $y_{ij} = 1$ if classifier $i$ correctly classifies observation $j$,  or assign $y_{ij} = 0$ otherwise. 
As we do not have the class labels or the ground truth of the observations, we cannot utilize this IRT ensemble for unsupervised anomaly detection.

In this paper we propose an IRT ensemble for unsupervised anomaly detection that uses its latent trait continuum to uncover the ground truth. This is different from the ensemble proposed by  \cite{Chen2020ensemble} because we use a different mapping of the IRT framework to ensemble learning. In their mapping the latent trait denotes the ability of the classifiers, whereas in our mapping the latent trait determines the hidden ground truth of observations. It is this different mapping that enables ensemble learning for unsupervised anomaly detection. Before examining different mappings, we discuss the problem of AD ensemble learning and introduce IRT briefly in Section~\ref{sec:background}. In Section~\ref{sec:method} we discuss different ways that IRT is mapped to machine learning problems and introduce our mapping for ensemble learning. Then, we compare different ensemble techniques on synthetic data in Section~\ref{sec:experiments} and using an extensive real data repository, conduct more comparisons in Section~\ref{sec:real}. Finally, we present our conclusions in Section~\ref{sec:conclusions}. As an additional contribution, we make these ensembles available in the R package \texttt{outlierensembles} \citep{outlierensembles}. 

\section{Problem Statement and Background}\label{sec:background}
\subsection{Unsupervised Anomaly Detection Ensembles}
\begin{figure}[!ht]
    \centering
    \includegraphics[scale=0.9]{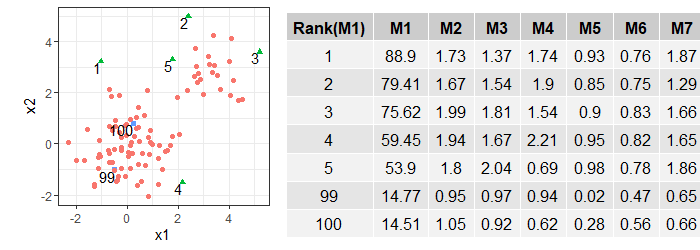}
    \caption{A synthetic example of 100 points and the resulting anomaly scores of 7 different AD methods denoted by M1 to M7. The table gives the anomaly scores of a set of selected points. These points are the top 5 and the bottom 2 anomalous points according to M1. These are marked in the figure by green triangles and blue rectangles respectively.}
    \label{fig:ADExample}
\end{figure}

Unsupervised Anomaly Detection (AD) methods generally give anomaly scores for each observation in the dataset with larger scores indicating anomalous observations. Thus, we have a ranking of observations in terms of anomalousness. Figure~\ref{fig:ADExample} shows an example of AD ensembling using a synthetic dataset of 100 points. The table shows the anomaly scores of 7 different methods, M1 to M7, on a selected set of 7 points -- the top 5 anomalous points and the bottom 2 anomalous points according to M1.  These points are denoted by green triangles and blue rectangles respectively. We see that the anomaly score ranks as well as the score ranges are different for each method. For example, M1 has much larger anomaly scores compared to the other methods. Also, M5 and M6 have scores between 0 and 1, while M2 and M3 have some scores greater than 1. The top ranked anomaly by M1 is ranked lower by M2 - M6. 

Given these heterogeneous methods and their anomaly scores, the task is to construct an ensemble score. As the anomaly scores of different methods have different scales, normalized AD scores are used to compute an ensemble score. Combination functions such as the average, the maximum or others  based on correlation and feature selection are typically used. We investigate the research problem of combining anomaly scores of multiple AD methods with the aim of achieving better performance compared to existing combination methods. To compare the performance of different combination functions, we use synthetic and real datasets with known labels or ground truth. Given that no single ensemble will outperform all other ensembles for all problem instances, we are interested in constructing an AD ensemble that is effective for a large proportion of the problem instances, surpassing the effectiveness of other ensembles. We state this formally as follows:

\begin{problemstatement} \textit{
Let $X$ be a dataset with $n$ observations and let $\bm{t}$ denote the ground truth vector of these observations with $\bm{t} \in \mathbb{R}^n$.  Let $\bm{u}_1, \bm{u}_2, \ldots, \bm{u}_\ell$ denote the anomaly scores of $\ell$ different AD methods on $X$ with $\bm{u}_i \in \mathbb{R}^n$ for $i \in \{1, \ldots, \ell\}$. Let $f_1, f_2, \ldots, f_k$ denote $k$ ensemble combination functions with $\bm{v}_1 = f_1 \left( \bm{u}_1, \ldots,  \bm{u}_\ell \right)$, \,\ldots, $\bm{v}_k = f_k \left( \bm{u}_1, \ldots,  \bm{u}_\ell \right)$ denoting their respective ensemble scores where $\bm{v}_j \in \mathbb{R}^n$ for $j \in \{1, \ldots, k\}$.  Let $\xi$ denote an AD performance metric such that $\xi\left(\bm{t}, \bm{v}_j\right) \in \mathbb{R}$. If  a combination function $f_b$ with  ensemble output $\bm{v}_b = f_b\left( \bm{u}_1, \ldots,  \bm{u}_\ell  \right) $ satisfies 
\begin{equation}\label{eq:probdef}
  \xi\left(\bm{t}, \bm{v}_b\right) > \xi\left(\bm{t}, \bm{v}_j\right)  \, , 
\end{equation}
for all $j \in \{1, \ldots, k\}, j \neq b$, then we say that $f_b$ is the best performing ensemble for $X$. For a collection of datasets  $\chi$, let $p_i$ be the proportion of datasets for which $f_i$ performs best according to equation~\eqref{eq:probdef}. Then, we seek an ensemble combination function $f_b$ for which 
$$ p_b  > p_j \, , $$
for all $b \neq j$ for any given collection of datasets $\chi$. 
}

\end{problemstatement} 

\subsubsection{The variables in the problem statement} 
We use a set of nearest neighbor based AD methods to build the ensemble: KNN-AGG \citep{Angiulli2002}, LOF \citep{Breuniq2000}, COF \citep{Tang2002} INFLO \citep{Jin2006}, KDEOS \citep{Schubert2014}, LDF \citep{Latecki2007} and LDOF \citep{Zhang2009}. We use the implementation in the R package \texttt{DDoutlier} \citep{ddoutlier} to compute the anomaly scores 
using these methods. These 7 methods either have a parameter $k$ or two parameters $k_{\min}$ and $k_{\max}$, which define the neighborhood for distance and density computations. We have used the default parameters  $k_{\min} = k=5$ and $k_{\max}= 10$ for all synthetic experiments. For experiments with real data, we have used two different parameter settings. 

As for ensemble methods, which are denoted by $f_1, \ldots, f_k$ in the \textit{Problem Statement}, we compare the performance of the following 7 ensembles: 
\begin{enumerate}
\item \textbf{IRT} ensemble is the proposed ensembling method, which we will discuss in detail in Section~\ref{sec:ensemble}.
\item \textbf{Average} ensemble computes the mean of the anomaly scores. We use the Average ensemble as a benchmark because the mean of the anomaly scores is often a useful indicator of anomalousness.  
\item \textbf{Greedy} ensemble \citep{Schubert2012} selects methods to include in the ensemble based on correlation values and requires a parameter $\kappa$, which is an estimate of the number of anomalies.  
\item \textbf{Averaged Greedy (Greedy-Avg)} ensemble runs the Greedy ensemble for different values of $k$ and computes the average. 
\item \textbf{Inverse Cluster Weighted Averaging (ICWA)} \citep{Chiang2017} is a cluster based ensembling method that uses affinity propagation \citep{BJ2007}.   
\item \textbf{Maximum Scoring (Max)} \citep{Aggarwal2015} uses the maximum of the anomaly scores for each observation. 
\item \textbf{Threshold Sum (Thresh)} \citep{Aggarwal2015} uses the summation of anomaly scores above the mean.
\end{enumerate} 
 
As our performance metric $\xi$, we use the area under the ROC curve denoted by AUC. For our dataset collections $\chi$, we use two collections from two separate approaches: \begin{inparaenum} \item synthetic data, for which we generate the anomalies using statistical methods and \item a data repository containing more than 12,000 AD datasets \citep{datasets}. 
\end{inparaenum}

Next, we briefly introduce IRT.


\subsection{IRT}
Item Response Theory (IRT) \citep{embretson2013item, van2013handbook}  refers to a family of latent trait models that is used to explain the relationship between unobservable characteristics such as intelligence or verbal ability and their observed outcomes such as responses to questionnaires. The underlying attributes such as intelligence or racial prejudice, which cannot be measured directly are modeled as latent variables. 

An IRT model is fitted using the responses of participants to test items, which are typically test questions. Depending on the type of response, different types of models are fitted. If the response is binary, a dichotomous IRT model is fitted. A  polytomous model is fitted to discrete valued responses and a continuous model to continuous valued responses. Furthermore, the term ``response'', which is traditionally used in the IRT literature can be somewhat misleading. For example, in educational testing a score derived from the participant's answers are used as responses in the IRT model. 

\subsubsection{Dichotomous IRT}\label{sec:dichotomous}
Suppose there are $i= 1, 2, \ldots, N$ participants and $j= 1, 2, \ldots, n$ test items. Let $y_{ij} \in \{0, 1\}$ denote the score or response of the $i^{\text{th}}$  participant to the $j^{\text{th}}$ test item. The discrimination parameter for test item $j$ is denoted by $\alpha_j$ and the difficulty parameter by $\beta_j$. These two parameters are used to build the 2-Parameter Logistic (2PL) model, giving the probability of a  correct response for each item given the ability level $\theta_i$ as 
\begin{equation}\label{eq:2plbasic}
     \Phi\left( y_{ij} =1 |\theta_i, \alpha_j, \beta_j \right) =   \frac{1}{1 + \exp\left(-\alpha_j \left( \theta_i - \beta_j \right) \right)} \, . 
 \end{equation}

\begin{figure}[ht]
    \begin{center}
    \centerline{\includegraphics[scale=0.5]{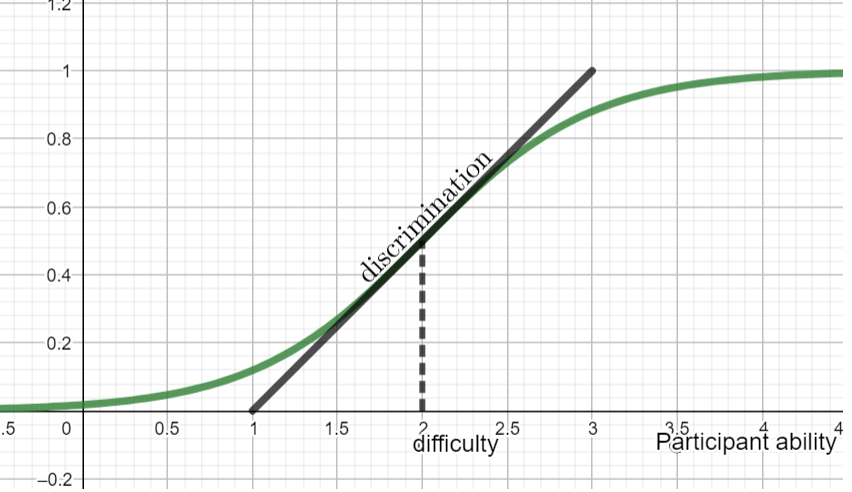}}
    \caption{The probability of the correct response for a test item using equation~\eqref{eq:2plbasic}. The discrimination $\alpha_j$ is proportional the slope of the tangent line at $\Phi = 0.5$.}
    \label{fig:2pl}
    \end{center}
\end{figure}

Figure~\ref{fig:2pl} shows an example item characteristic curve fitted using a 2PL model. The discrimination of this test item $\alpha_j$ is proportional to the slope of the tangent line drawn at $\Phi = 0.5$. When $\theta_i = \beta_j$, $\Phi = 0.5$, that is when the difficulty of the test question equals the ability of the participant, the probability of obtaining the correct answer is 0.5. If the participant ability is higher than the item difficulty, then the probability of getting the correct answer is higher than 0.5 and vice versa. 

\subsubsection{Continuous IRT}\label{sec:continuous}
\begin{figure}[ht]
    \begin{center}
    \centerline{\includegraphics[scale=0.75]{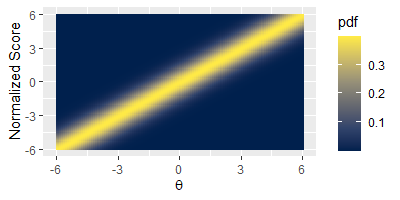}}
    \caption{The probability of the normalized response $z$ for test item $j$ using equation~\eqref{eq:CRM4} with $\alpha_j=1$, $\beta_j=0$ and $\gamma_j=1$. }
    \label{fig:ctsIRT}
    \end{center}
\end{figure}

 Next, we consider the continuous response model formulation proposed by Wang and Shojima \citep{Wang1998, SHOJIMA2005}, which extends the continuous model introduced by Samejima \citep{samejima1969estimation}. Suppose there are $i= 1, 2, \ldots, N$ participants and $j= 1, 2, \ldots, n$ test items. As before, let $y_{ij}$ denote the score or response of the $i^{\text{th}}$  participant to the $j^{\text{th}}$ test item. Then we scale $y_{ij} \in (0, k)$ by defining  $x_{ij} = y_{ij}/k$, so that $x_{ij} \in (0, 1)$. Let $\theta_i$ denote the trait variable of the $i^{\text{th}}$ participant and $\bm{\lambda}_j = (\alpha_j, \beta_j, \gamma_j)^T$ denote the item parameters of the $j^{\text{th}}$ test item where $\alpha_j$ denotes the discrimination, $\beta_j$  difficulty and $\gamma_j$ a scaling coefficient. Using the Gaussian,  the probability of a participant with ability $\theta$  obtaining a score $x$ or higher for item $j$ is modeled as
\begin{equation}\label{eq:CRM1}
P\left( X \geq x |\theta \right) = \frac{1}{\sqrt{2\pi}} \int_{-\infty}^{v} e^{-\frac{t^2}{2}} dt \, , 
\end{equation}
where
\begin{equation}\label{eq:CRM2}
    v = \alpha_j\left(\theta - \beta_j - \gamma_j \ln \frac{x}{1 - x} \right)\, . 
\end{equation}
With the reparametrization  
\begin{equation}\label{eq:CRM3}
    z = \ln \frac{x}{1 - x} \, , 
\end{equation}
the probability density function $f\left(z|\theta \right)$ obtained by differentiating the cumulative density function $ 1- P\left( X \geq x |\theta \right) $ using equation~\eqref{eq:CRM1} is given by 
\begin{equation}\label{eq:CRM4}
    f\left(z|\theta \right) = \frac{\alpha_j\gamma_j}{\sqrt{2\pi}} \exp \left( -\frac{\alpha_j^2}{2} \left( \theta - \beta_j - \gamma_j z \right)^2 \right) \, .
\end{equation}
Equation~\eqref{eq:CRM4} prescribes a surface of probabilities for every test item $j$ as shown in Figure~\ref{fig:ctsIRT}. That is, $ f\left(z|\theta \right) $ is a function of the normalized scores $z$ and the latent trait $\theta$ for fixed values of $\alpha_j$, $\beta_j$ and $\gamma_j$.

Both test items in Figures~\ref{fig:2pl} and~\ref{fig:ctsIRT} have positive discrimination. That is, as the participant ability $\theta$ increases, the probability of getting a higher score increases.  If an item has negative discrimination, the slope of the line in Figure~\ref{fig:2pl} would be negative and similarly, the high density yellow region in Figure~\ref{fig:ctsIRT} will have a negative slope. In educational testing, items with negative discrimination are considered non-value adding because participants with higher ability obtain lower scores than  participants with lower ability for such items. As such, these items are removed before the model is fitted. We have extended the present educational testing model  \citep{SHOJIMA2005} to include negative discrimination items  \citep{skksm2020} as our focus is on algorithms and their performance and as such we do not want to remove negative discrimination items from the pool. 

The  continuous IRT model \citep{SHOJIMA2005} maximizes the log-likelihood 
\begin{align}\label{eq:CRM12-1}
    E  & = N \sum_{j=1}^n \left(\ln \alpha_j  + \ln\gamma_j \right) \notag \\
    & - \frac{1}{2} \sum_{i=1}^N \sum_{j=1}^n \alpha_j^2 \left( \left(  \beta_j + \gamma_j z_{ij} - \mu_i^{(t)} \right)^2  + \sigma^{(t)2} \right) \notag \\
    & + \ln p\left(\bm\Lambda \right) + \text{const} \, ,
\end{align}
where $\alpha_j, \gamma_j >0$ for all $j$,    $\mu_i^{(t)}$ and $\sigma^{(t)}$ denote the mean and standard deviation of $\theta_i$ computed in the $t^{\text{th}}$ iteration,   $\bm{\Lambda} = \left(\bm{\lambda_1}, \ldots, \bm{\lambda_n} \right)$ with $\bm{\lambda_j} = \left(\alpha_j, \beta_j, \gamma_j \right)^T$ and $p$ denotes a prior distribution. While fitting the IRT model if $\alpha_j$ or $\gamma_j$ becomes negative, then it stops the convergence process as equation~\eqref{eq:CRM12-1} becomes undefined due to the logarithm. However, the probability density function given in equation~\eqref{eq:CRM4} is well defined if both $\alpha_j$ and $\gamma_j$ are negative. Thus, equation~\eqref{eq:CRM12-1} only accommodates half of the possible $\alpha_j$ and $\gamma_j$ values, because the real constraint is that the product $\alpha_j \gamma_j >0$. We correct this problem by updating the log-likelihood function as follows:
\begin{align}\label{eq:CRM12}
    E  & = N \sum_{j=1}^n \left(\ln |\alpha_j|  + \ln|\gamma_j| \right) \notag \\
    & - \frac{1}{2} \sum_{i=1}^N \sum_{j=1}^n \alpha_j^2 \left( \left(  \beta_j + \gamma_j z_{ij} - \mu_i^{(t)} \right)^2  + \sigma^{(t)2} \right) \notag \\
    & + \ln p\left(\bm\Lambda \right) + \text{const} \, .
\end{align}
 
This modified function agrees with the real constraint $\alpha_j \gamma_j >0$ and involves minor updates to the parameter values. Following through the computation \citep{SHOJIMA2005}, the discrimination parameter $\alpha_j$ changes to
 \begin{equation}\label{eq:CRM14}
      \alpha_j^{(t+1)}  =  \sign\left(\gamma_j^{(t+1)}\right) \alpha_{j, \text{ori}}^{(t+1)}\, ,    
\end{equation}
 where $\alpha_{j, \text{ori}}$ denotes the discrimination parameter in the original computation and $t$ denotes the iteration. The parameter $\gamma_j$, which is a scaling coefficient can obtain negative values in this formulation, while the other parameters remain unchanged. More details of this update are available at \cite{skksm2020}. 
 
The maximum likelihood estimate of the latent trait parameter $\theta_i$ is computed using the converged estimates of the item parameters $\hat{\alpha}_j$, $\hat{\beta}_j$ and $\hat{\gamma}_j$ as follows:
 \begin{equation}\label{eq:CRM15}
     \theta_i = \frac{\sum_j \hat{\alpha}_j^2 \left(\hat{\beta}_j + \hat{\gamma}_j z_{ij} \right)}{  \sum_j \hat{\alpha}_j^2} \, .
 \end{equation}
We see that the latent trait parameter $\theta_i$, which denotes the ability of participant  $i$ is a function of  item discrimination $\alpha_j$,  difficulty $\beta_j$, scaling factor $\gamma_j$ as well as the test scores $z_{ij}$. 

\section{Mapping IRT to ensemble learning}\label{sec:method}

We start this Section by discussing the traditional IRT mapping used in educational testing and psychometrics. Then we introduce newer IRT mappings currently used for machine learning tasks including our proposed IRT mapping for ensemble learning.

\subsection{Educational testing and psychometrics}\label{sec:educational}
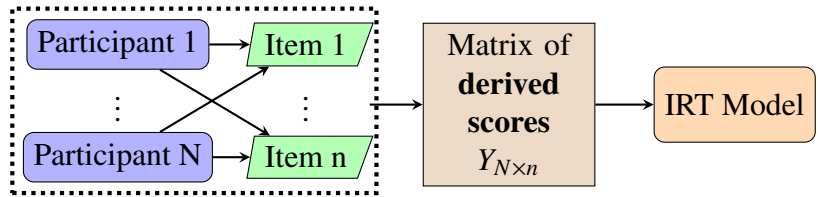
\begin{figure}[!ht]
    \centering
    \begin{tikzpicture}
    \node (p1) [person] {Participant 1};
    \node (p2) [person, below of=p1, yshift=-0.5cm] {Participant N};
    \node (t1) [testitem, right of=p1, xshift=1.5cm] {Item 1};
    \node (t2) [testitem, right of=p2, xshift=1.5cm] {Item n};
    \path (p1) -- node[auto=false]{\vdots} (p2);
    \path (t1) -- node[auto=false]{\vdots} (t2);
    \node (invisible1) [below of=t1, xshift=0.7cm, yshift=0.2cm]{};
    \node (m1) [object, right of=invisible1, xshift=1cm] {Matrix of  \textbf{derived scores} $Y_{N \times n}$};
    \node (m2) [model, right of=m1, xshift=2cm] {IRT Model};
     \draw [arrow] (p1) -- (t1);
     \draw [arrow] (p1) -- (t2);
     \draw [arrow] (p2) -- (t1);
     \draw [arrow] (p2) -- (t2);
      \node [fit=(p1) (p2) (t1) (t2),draw,dotted,ultra thick,black] {};
     \draw [arrow] (invisible1) -- (m1);
      \draw [arrow] (m1) -- (m2);
    \end{tikzpicture}
    \caption{Participants doing test items in the educational testing scenario. The matrix of derived test scores/marks are used to fit the IRT model.}
    \label{fig:IRTEdu}
\end{figure}

In educational testing participants sit for tests, and the test item scores are used to fit the IRT model. In this setting $y_{ij}$ is the score of the $i^{\text{th}}$ participant to the $j^{\text{th}}$ test item.  This is shown in Figure~\ref{fig:IRTEdu} with the forward arrow indicating the action of doing each test item. From this IRT model we obtain  item discrimination and difficulty for test items, and and participant ability for participants. 

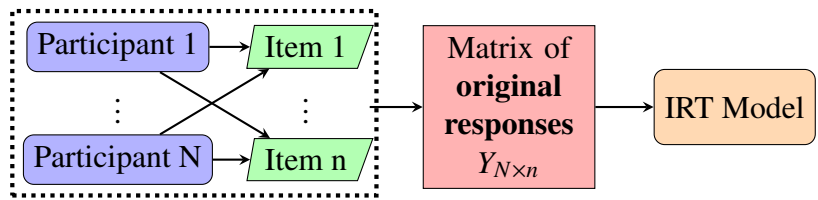
\begin{figure}[!ht]
    \centering
    \begin{tikzpicture}
    \node (p1) [person] {Participant 1};
    \node (p2) [person, below of=p1, yshift=-0.5cm] {Participant N};
    \node (t1) [testitem, right of=p1, xshift=1.5cm] {Item 1};
    \node (t2) [testitem, right of=p2, xshift=1.5cm] {Item n};
    \path (p1) -- node[auto=false]{\vdots} (p2);
    \path (t1) -- node[auto=false]{\vdots} (t2);
    \node (invisible1) [below of=t1, xshift=0.7cm, yshift=0.2cm]{};
    \node (m1) [object, right of=invisible1, xshift=1cm,  fill=red!30] {Matrix of  \textbf{original responses} $Y_{N \times n}$};
    \node (m2) [model, right of=m1, xshift=2cm] {IRT Model};
     \draw [arrow] (p1) -- (t1);
     \draw [arrow] (p1) -- (t2);
     \draw [arrow] (p2) -- (t1);
     \draw [arrow] (p2) -- (t2);
      \node [fit=(p1) (p2) (t1) (t2),draw,dotted,ultra thick,black] {};
     \draw [arrow] (invisible1) -- (m1);
      \draw [arrow] (m1) -- (m2);
    \end{tikzpicture}
    \caption{Participants doing a psychometrics questionnaire. The matrix of original responses are used to fit the IRT model.}
    \label{fig:IRTSelf}
\end{figure}

In educational testing, the test item scores or marks that are derived from the original user responses are used in the IRT model. In some branches of psychometrics such as self esteem or personality assessment studies, the original responses are used in the IRT model \citep{Gray-Little1997}. This is because in such scenarios the test items are statements such as \textit{I feel that I have a number of good qualities} or \textit{I take a positive attitude toward myself} \citep{rosenberg1965rosenberg}, and thus requires the user to agree or disagree with the statements using a rating scale. In such instances there is nothing to derive in terms of the responses, i.e. there is no correct or wrong answer for any question. Therefore, the original responses are used in the IRT model. Similarly, the original responses are also used in studies investigating underlying prejudice. \cite{Villano2017} use IRT to explore stereotyping Roma people or gypsies in Italy. They used surveys having statements such as \textit{Roma people do not want to integrate and prefer to be marginalized.} Again, the original responses are used in these IRT models.  Figure~\ref{fig:IRTSelf} shows this model. The latent trait in this IRT model determines a hidden quality such as self esteem, a hidden prejudice or some personality trait of the participants.

Here, we make a distinction between the original responses and the derived scores. Derived scores are obtained using an accuracy measure on the original responses. Traditional IRT uses both; educational testing uses derived scores which is an accuracy measure while personality trait studies uses original responses. 


\subsection{Algorithm evaluation}\label{sec:algoeval}
\begin{figure}[!ht]
    \centering
    \begin{tikzpicture}
    \node (p1) [person] {Algorithm 1};
    \node (p2) [person, below of=p1, yshift=-0.5cm] {Algorithm N};
    \node (t1) [testitem, right of=p1, xshift=2.5cm] {Observation 1};
    \node (t2) [testitem, right of=p2, xshift=2.5cm] {Observation n};
    \path (p1) -- node[auto=false]{\vdots} (p2);
    \path (t1) -- node[auto=false]{\vdots} (t2);
    \node (invisible1) [below of=t1, xshift=1.2cm, yshift=0.2cm]{};
       \node (m1) [object, right of=invisible1, xshift=1cm] {Matrix of  \textbf{derived accuracies} $Y_{N \times n}$};
    \node (m2) [model, right of=m1, xshift=2cm] {IRT Model};
     \draw [arrow] (p1) -- (t1);
     \draw [arrow] (p1) -- (t2);
     \draw [arrow] (p2) -- (t1);
     \draw [arrow] (p2) -- (t2);
      \node [fit=(p1) (p2) (t1) (t2),draw,dotted,ultra thick,black] {};
     \draw [arrow] (invisible1) -- (m1);
      \draw [arrow] (m1) -- (m2);
     \end{tikzpicture}
    \caption{Algorithms acting on dataset observations in the standard algorithm evaluation setting. An accuracy measure is used in  matrix $Y_{N \times n}$. }
    \label{fig:IRTAlgo}
\end{figure}
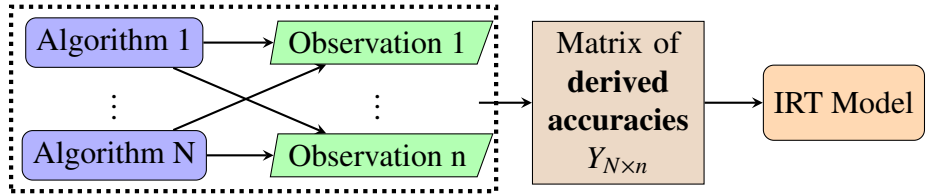

In algorithm evaluation the standard approach is to map the participants to algorithms, and test items to dataset observations \citep{Martinez-Plumed2019, Chen2020}. This is illustrated in Figure~\ref{fig:IRTAlgo}. In this setting $y_{ij}$ denotes an accuracy measure such as  classification  accuracy of the $i^{\text{th}}$ algorithm on the $j^{\text{th}}$ observation making the rows of matrix $Y$ depict the algorithms and the columns, observations.  The resulting IRT model gives the ability of the algorithms and the difficulty and discrimination of the observations.  

We used a different, inverted mapping that uses datasets instead of observations to evaluate algorithm performance \citep{skksm2020}. In this mapping participants were mapped to datasets and test items were mapped to algorithms. We used an accuracy measure depending on the class of algorithms. In this setting $y_{ij}$ denotes the accuracy of dataset $i$ to algorithm $j$ making the rows of matrix $Y$ give the datasets and the columns, algorithms. The resulting IRT model gives the discrimination and difficulty of the algorithms and the ability parameter for the datasets, which in essence denotes dataset easiness/difficulty. 

\begin{figure}[!ht]
    \centering
    \begin{tikzpicture}
    \node (p1) [person] {Dataset 1};
    \node (p2) [person, below of=p1, yshift=-0.5cm] {Dataset N};
    \node (t1) [testitem, right of=p1, xshift=2.5cm] {Algorithm 1};
    \node (t2) [testitem, right of=p2, xshift=2.5cm] {Algorithm n};
    \path (p1) -- node[auto=false]{\vdots} (p2);
    \path (t1) -- node[auto=false]{\vdots} (t2);
    \node (invisible1) [below of=t1, xshift=1cm, yshift=0.2cm]{};
       \node (m1) [object, right of=invisible1, xshift=1cm] {Matrix of  \textbf{derived accuracies} $Y_{N \times n}$};
    \node (m2) [model, right of=m1, xshift=2cm] {IRT Model};
     \draw [arrow] (p1) -- (t1);
     \draw [arrow] (p1) -- (t2);
     \draw [arrow] (p2) -- (t1);
     \draw [arrow] (p2) -- (t2);
      \node [fit=(p1) (p2) (t1) (t2),draw,dotted,ultra thick,black] {};
     \draw [arrow] (invisible1) -- (m1);
      \draw [arrow] (m1) -- (m2);
     \end{tikzpicture}
    \caption{Datasets acting on algorithms in the algorithm evaluation setting. An accuracy measure is used in the IRT model.}
    \label{fig:Invertedmapping1}
\end{figure}
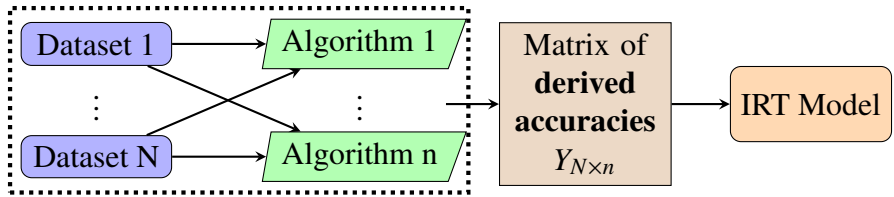


\subsection{Chen's classification ensemble}
\cite{Chen2020ensemble} use the mapping shown in Figure~\ref{fig:IRTAlgo} for their classification ensemble. They use  classification accuracy in their model, that is $y_{ij} = 1$ if classifier $i$  correctly classifies observation $j$, or $y_{ij} = 0$ otherwise. From this mapping they obtain the algorithm ability parameter $\theta_i$ and compute a weight for each algorithm as
\begin{equation}
    w_i = \frac{e^{\theta_i}}{\sum_{i=1}^N e^{\theta_i} } \, .
\end{equation}
The weights $w_i$ are computed from a training set and applied to a test set to obtain the ensemble output as a weighted sum. 

This ensemble framework cannot be employed for unsupervised algorithms because it uses an accuracy measure, which is computed by comparing the algorithm output to the ground truth. When an accuracy measure is used in the IRT model, the parameter $\bm{\theta} = \{\theta_i\}_{i=1}^N$ denotes the ability of the algorithm/participant in the standard sense of the word, that is, it is indeed a skill.  When the original responses are used, $\bm{\theta}$ denotes a hidden quality such as self-esteem, a prejudice or a personality trait.

\subsection{Proposed AD ensemble}\label{sec:ensemble}

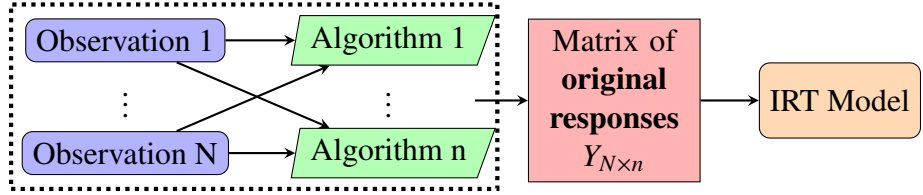
\begin{figure}[!ht]
    \centering
    \begin{tikzpicture}
    \node (p1) [person] {Observation 1};
    \node (p2) [person, below of=p1, yshift=-0.5cm] {Observation N};
    \node (t1) [testitem, right of=p1, xshift=2.5cm] {Algorithm 1};
    \node (t2) [testitem, right of=p2, xshift=2.5cm] {Algorithm n};
    \path (p1) -- node[auto=false]{\vdots} (p2);
    \path (t1) -- node[auto=false]{\vdots} (t2);
    \node (invisible1) [below of=t1, xshift=1cm, yshift=0.2cm]{};
     \node (m1) [object, right of=invisible1, xshift=1cm, fill=red!30] {Matrix of  \textbf{original responses} $Y_{N \times n}$};
    \node (m2) [model, right of=m1, xshift=2cm] {IRT Model};
     \draw [arrow] (p1) -- (t1);
     \draw [arrow] (p1) -- (t2);
     \draw [arrow] (p2) -- (t1);
     \draw [arrow] (p2) -- (t2);
      \node [fit=(p1) (p2) (t1) (t2),draw,dotted,ultra thick,black] {};
     \draw [arrow] (invisible1) -- (m1);
      \draw [arrow] (m1) -- (m2);
    \end{tikzpicture}
    \caption{The proposed IRT mapping for AD ensemble learning. Observations acting on algorithms with original responses used in matrix $Y_{N \times n}$.}
    \label{fig:IRTEnsemble2}
\end{figure}

We propose the following IRT mapping for unsupervised AD ensemble learning. 

\begin{enumerate}
   \item We use the original responses instead of an accuracy measure. Specifically, we use the standardized original responses of algorithms. The standardization is done to  effectively compare the output of different algorithms. Especially in anomaly detection, the anomaly responses of different algorithms may be vastly different (see Figure~\ref{fig:ADExample}) and as such standardizing the output of each algorithm to the interval $[0,1]$ places all algorithms on the same scale.
   
    \item We map participants to dataset observations and test items to algorithms. 
    
\end{enumerate}  This mapping is shown in Figure~\ref{fig:IRTEnsemble2}. In this setting $y_{ij}$ denotes the scaled response of observation $i$ to algorithm $j$. This is different from the algorithm evaluation mapping in Figure~\ref{fig:Invertedmapping1}, because it uses the original responses instead of accuracy measures and also it is constructed at an observational level instead of at a dataset level. Furthermore, the goal of the mapping in Figure~\ref{fig:Invertedmapping1} is to evaluate a portfolio of algorithms using an accuracy measure, whereas here we are interested in constructing an ensemble using the latent trait continuum.  

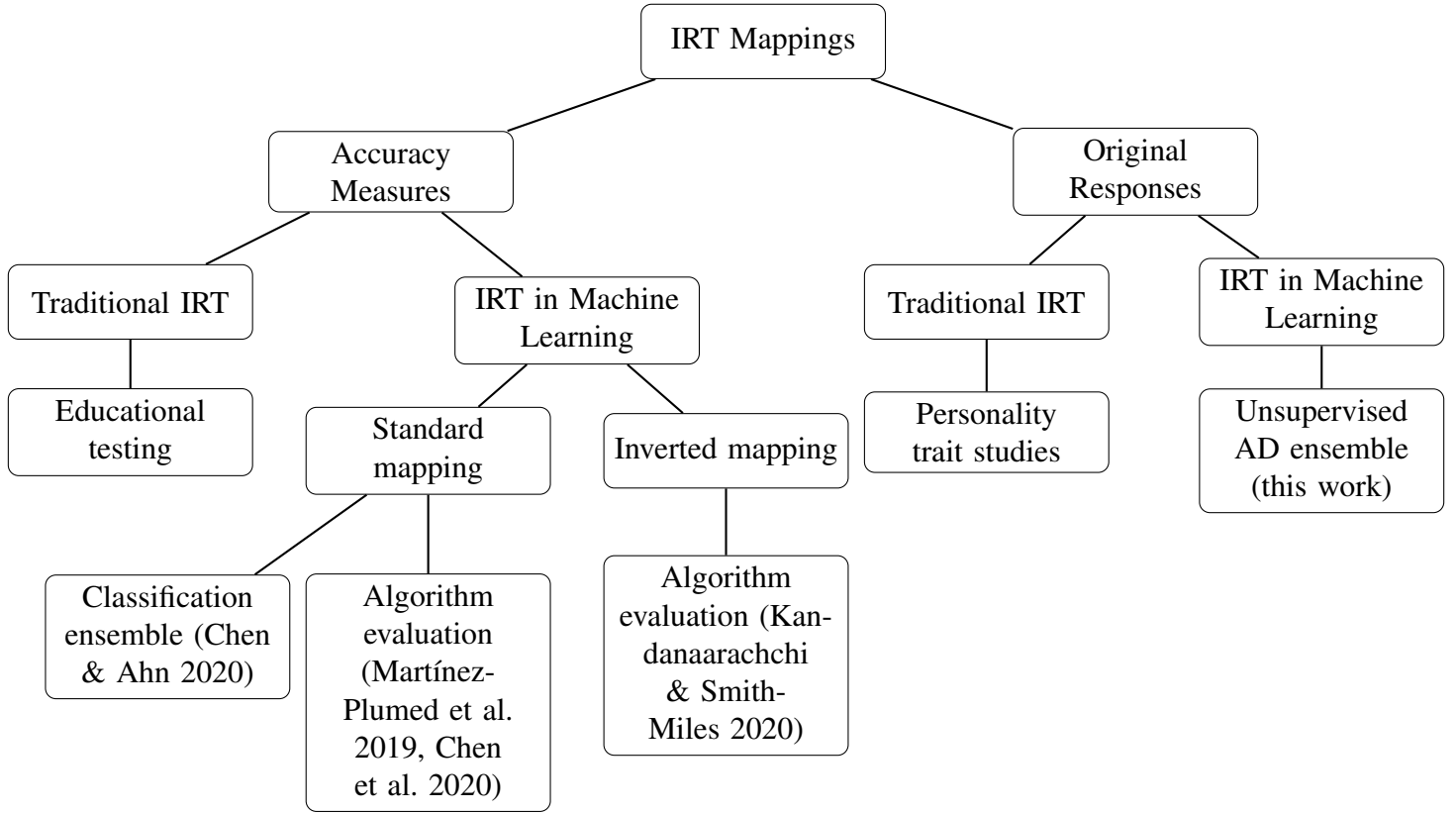
\begin{figure}[!ht]
    \centering
    \begin{tikzpicture}
    \node (p1) [std] {IRT Mappings};
    \node(l1) [std, below of=p1, yshift=-0.75cm, xshift=-5cm] {Accuracy Measures};
    \node(r1) [std, below of=p1, yshift=-0.75cm, xshift=5cm] {Original Responses};
    \node(ll2) [std, below of=l1, yshift=-0.75cm, xshift=-3.5cm] {Traditional IRT};
    \node(ll3) [std, below of=ll2, yshift=-0.75cm] {Educational testing};
    \node(lr2) [std, below of=l1, yshift=-1cm, xshift=2.5cm] { IRT in Machine Learning};
    \node(lrl3) [std, below of=lr2, yshift=-0.75cm, xshift=-2cm] { Standard mapping}; 
    \node(lrr3) [std, below of=lr2, yshift=-0.75cm, xshift=2cm] { Inverted mapping}; 
    \node(lrl4) [std, below of=lrl3, yshift=-2.25cm] {Algorithm evaluation \citep{Martinez-Plumed2019, Chen2020}};
     \node(lrr4) [std, below of=lrl3, yshift=-1.5cm, xshift=-3.5cm] {Classification ensemble \citep{Chen2020ensemble}};
    \node(lrrl4) [std, below of=lrr3, yshift=-1.75cm] { Algorithm evaluation \citep{skksm2020}};     
    \node(rl2) [std, below of=r1, yshift=-0.75cm, xshift=-2cm] {Traditional IRT};
     \node(rr2) [std, below of=r1, yshift=-0.75cm, xshift=2.5cm] { IRT in Machine Learning};
    \node(rll3) [std, below of=rl2, yshift=-0.75cm] {Personality trait studies};
    \node(rrr3) [std, below of=rr2, yshift=-1cm] {Unsupervised AD ensemble (this work)};
     \draw [line] (p1) -- (l1);
     \draw [line] (l1) -- (ll2); 
     \draw [line] (l1) -- (lr2);
     \draw [line] (ll2) -- (ll3);
     \draw [line] (lr2) -- (lrl3);
     \draw [line] (lr2) -- (lrr3);
     \draw [line] (lrl3) -- (lrr4);
     \draw [line] (lrl3) -- (lrl4);
     \draw [line] (lrr3) -- (lrrl4);
     
     \draw [line] (p1) -- (r1);
     \draw [line] (r1) -- (rl2);
     \draw [line] (r1) -- (rr2);
     \draw [line] (rl2) -- (rll3);
     \draw [line] (rr2) -- (rrr3);
     \end{tikzpicture}
    \caption{Different IRT mappings in traditional and machine learning settings.}
    \label{fig:differentIRTmappings}
\end{figure}

We summarize the different IRT configurations or mappings in Figure~\ref{fig:differentIRTmappings}. Then we examine the IRT parameters of the AD ensemble framework.  
\subsubsection{The latent trait }
In educational testing, the latent trait continuum holds participants ordered by their ability. As test item scores are used to fit the IRT model, the ability generally increases with the test item scores. However, ability is not perfectly correlated with the total score, because the hardness of the test items is taken into account. For example a participant who obtains correct answers for more difficult test items and misses a couple of easy test questions will have a higher ability than a participant who misses the harder questions but gets all the easy test questions correct.

When the original responses are used in the IRT model, the latent trait continuum holds a hidden quality such as self esteem, which is tested by the survey. Again, the participants are ordered in the latent trait continuum by this quality. When participants are mapped to observations in the ensemble setting, the parameter $\bm{\theta}$ denotes a certain quality about the observations. In addition, when the standardized algorithm response is used instead of an accuracy measure, the new trait parameter increases with the algorithm response. But the algorithm response is the anomaly score assigned to each observation.  As such, $\bm{\theta}$  increases with the anomaly score, which is a measure of the anomalousness of the observations.  Thus, the parameter $\theta_i$ denotes the level of anomalousness of observation $i$, taking into account all the AD methods. Therefore, it is an ensemble score.



\subsubsection{The difficulty parameter $\beta$}
The interpretation of the difficulty parameter does not change much. For positively discriminated algorithms, if an observation $i$ has $\theta_i > \beta_j$, then it makes the normalized response $z > 0 $ more probable. Therefore, $\beta_j$ is the \textit{anomalousness threshold} of algorithm $j$ for obtaining $z=0$. For anomaly detection a higher anomalousness threshold $\beta$ is preferred as this would result in fewer anomalies. However, we need to consider the discrimination parameter as well to get a better understanding of the algorithm. 
\begin{figure}[ht]
    \begin{center}
    \centerline{\includegraphics[scale=0.8]{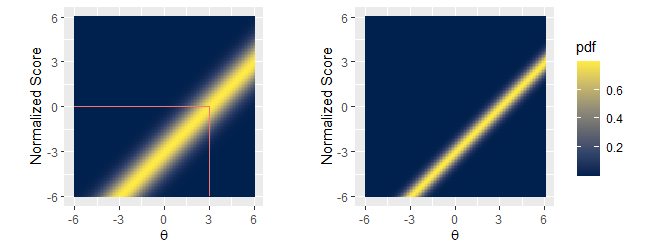}}
    \caption{The probability $f(z|\theta)$ for test item $j$ using equation~\eqref{eq:CRM4} with $\alpha_j=1$, $\beta_j=3$ and $\gamma_j=1$ on the left and $\alpha_j=2$, $\beta_j=3$ and $\gamma_j=1$ on the right. }
    \label{fig:difficultyDiscrimination}
    \end{center}
\end{figure}

\subsubsection{The discrimination parameter $\alpha$}
Similar to the difficulty parameter, the interpretation of the discrimination parameter remains unchanged with higher values of $\alpha$ denoting sharper transitions in probability for different values of $z$ or $\theta$. This is illustrated in Figure~\ref{fig:difficultyDiscrimination}. The plot on the left has $\alpha_j = 1$ and the plot on the right has $\alpha_j = 2$. Both have $\beta_j = 3$. An anomaly detection method with a  lower discrimination parameter would give similar scores for anomalies and non-anomalies. Therefore, a higher discrimination parameter is preferred.

\subsubsection{The scaling parameter $\gamma$}
The parameter $\gamma$ is not a traditional IRT parameter. Instead it is used for scaling purposes to assure that $f(z|\theta)$ is indeed a probability density function, i.e. the area under the curve for any fixed $\theta$ adds up to 1.  Even though it does not have a traditional IRT meaning, the scaling parameter $\gamma_j$ plays a part in determining the latent trait spectrum. Going back to equation~\eqref{eq:CRM15} we see that the latent trait parameter,
\begin{align}\label{eq:CRM152}
     \theta_i & = \frac{\sum_j \hat{\alpha}_j^2 \left(\hat{\beta}_j + \hat{\gamma}_j z_{ij} \right)}{  \sum_j \hat{\alpha}_j^2} \, , \notag \\
     & =  \sum_j\left( \frac{\hat{\alpha}_j^2\hat{\beta}_j}{\sum_j \hat{\alpha}_j^2}   +  \frac{\hat{\alpha}_j^2 \hat{\gamma}_j }{\sum_j \hat{\alpha}_j^2} z_{ij} \right) \, , \notag  \\
     & = \sum_j\left( \zeta_j + \omega_j z_{ij}  \right) \, , 
\end{align}
with $\zeta_j = \frac{\hat{\alpha}_j^2\hat{\beta}_j}{\sum_j \hat{\alpha}_j^2}  $ and $\omega_j = \frac{\hat{\alpha}_j^2 \hat{\gamma}_j }{\sum_j \hat{\alpha}_j^2} $. Noting that $\sum_j \zeta_j$ and $\sum_j \hat{\alpha}_j^2 $ are constant for a given IRT model, we see that the anomaly response of the $i^{\text{th}}$ observation to the $j^{\text{th}}$ algorithm, $z_{ij}$ gets weighted by $\omega_j$, which is a scaled product of the discrimination  and the scaling parameters $\alpha_j$ and $\gamma_j$ of AD algorithm $j$. Therefore, if an AD method has high discrimination and scaling parameters, the anomaly responses of that method are weighted higher and contribute more to the ensemble score.

\section{Experiments with synthetic data}\label{sec:experiments}
First we discuss the parameters of the ensemble methods. The Greedy ensemble requires a parameter $\kappa$, which is an estimate of the number of anomalies in the dataset. In our synthetic experiments, we have set $\kappa$ to  the exact number of anomalies so that we evaluate the Greedy ensemble at its best. However, as we do not know the number of anomalies in practice, we consider the averaged Greedy ensemble, which runs the Greedy ensemble for different values of $\kappa$ and averages the output. For the synthetic examples we vary $\kappa$ from 1 to 10. The clustering based ensemble method ICWA can be used with various similarity measures. Of the similarity measures listed in \cite{Chiang2017}, we use Pearson's correlation. 

Using these parameters, we illustrate the performance of the ensembles using a simple example in $\mathbb{R}^2$. Then we conduct 3 iterated experiments,  which are set up so that the anomalies become more marked with each iteration. We compare the performance of the ensemble methods for these iterated experiments. 



\subsection{Example}\label{sec:example}
\begin{figure}[ht]
    \begin{center}
    \centerline{\includegraphics[scale=0.8]{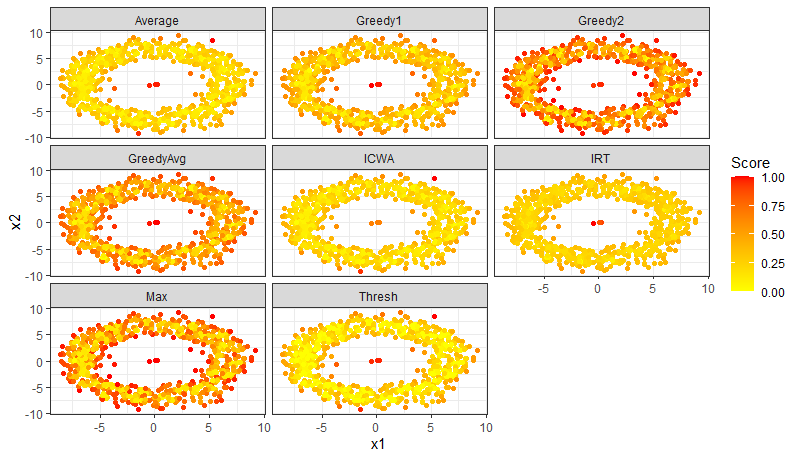}}
    \caption{The AD ensemble scores for the Example. Annulus with 3 anomalies at the center. }
    \label{fig:Eample}
    \end{center}
\end{figure}

We consider a simple example of an annulus with 3 anomalies placed at the center. We obtain the anomaly responses from the AD methods and construct the AD ensembles. Figure~\ref{fig:Eample} shows the scaled ensemble scores of the 7 ensemble methods. The Greedy ensemble is used twice with different parameters, $\kappa =3$ and $\kappa =10$.  
 
The Average ensemble performs reasonably well obtaining high ensemble scores for the anomalies in the center. However, it also obtains high scores for at least 2 more points on the boundary of the annulus. The Greedy ensemble with $\kappa=3$, labeled \textit{Greedy1}, also gives good performance. But by setting $\kappa=3$, we have selected the best value for parameter $\kappa$ as there are exactly 3 anomalies. When we use $\kappa = 10$, the Greedy ensemble performs poorly as seen in the pane labeled \textit{Greedy2}. Thus, the Greedy ensemble is very sensitive to parameter selection. As Greedy-Avg averages the ensemble scores for different values of $\kappa$, its performance lies in between \textit{Greedy1} and \textit{Greedy2}. ICWA finds two boundary points as most anomalous, disregarding the real anomalies. IRT gives high ensemble scores to the anomalies and low scores to the other points. The Max ensemble gives poor performance, while Thresh is similar to the Average ensemble.  

A point of interest is that the IRT ensemble gives a low ensemble score to the observation positioned approximately at $(5, 7.5)$, while Average, Thresh and ICWA gives relatively high scores to that observation. Upon inspection we see that the scaled anomaly responses for KNN-AGG, LOF, COF, INFLO, KDEOS, LDF and LDOF are 0.48, 0.56, 0.62, 1, 0.98, 0.74 and 1 respectively, where all anomaly responses are scaled to the interval $[0,1]$.  Thus, INFLO, KDEOS and LDOF have very high anomaly responses for this observation. But, as seen from Table~\ref{tab:IRTParas} the discrimination parameter $\alpha_j$ is small for LDOF, INFLO and KDEOS compared with the $\alpha_j$ values for LOF and LDF. As a result, these responses get discounted when computing the latent score. This illustrates how the IRT ensemble downplays certain methods while highlighting other methods.


\begin{table}[!ht]
	\centering
	\caption{IRT parameters discrimination $(\alpha)$,  difficulty $(\beta)$ and the scaling parameter $(\gamma)$ for the example and the synthetic experiments.}
	\footnotesize
	\begin{tabular}{lccc|ccc|ccc|ccc}
		\toprule
    \multirow{2}{*}{Method} &  \multicolumn{3}{c}{Example} &  \multicolumn{3}{c}{EX1} & \multicolumn{3}{c}{EX2} & \multicolumn{3}{c}{EX3}	\\	
  \cmidrule{2-13} & $\alpha$ & $\beta$ & $\gamma$ &   $\alpha$ & $\beta$ & $\gamma$ &  $\alpha$ & $\beta$ & $\gamma$ & $\alpha$ & $\beta$ & $\gamma$ \\
        \midrule
    KNN-AGG & 1.6 & 3.3 & 0.7 & 1.4 & 2.6 & 0.7  & 1.3 & 2.1 & 0.7 & 2.1 & 1.8 & 0.8\\ 
    LOF    & 14.0 & 3.3 & 0.9 & 2.7 & 1.7 & 0.9  & 5.0 & 2.3 & 0.8 & 2.3 & 1.9 & 0.9 \\ 
    COF     & 0.9 & 2.7 & 0.5 & 1.1 & 1.1 & 0.7  & 1.0 & 2.3 & 0.6 & 1.2 & 1.5 & 0.6\\ 
    INFLO   & 0.3 & 7.7 & 0.1 & 0.6 & 2.6 & 0.4  & 0.5 & 3.7 & 0.3 & 0.7 & 3.2 & 0.4\\ 
    KDEOS   & 1.1 & 0.0 & 1.3 & 0.9 & 0.0 & 1.2  & 0.8 & 0.0 & 1.2 & 1.0 & 0.0 & 1.3\\
    LDF     & 6.2 & 2.4 & 0.9 & 1.4 & 1.0 & 0.9  & 2.1 & 1.1 & 0.8 & 1.6 & 0.6 & 0.9\\
    LDOF    & 0.6 & 5.8 & 0.4 & 0.8 & 2.4 & 0.6  & 0.6 & 4.1 & 0.5 & 0.9 & 2.3 & 0.6\\
		 \bottomrule
	\end{tabular}
	\label{tab:IRTParas}
\end{table}

\subsection{Experiment 1}\label{sec:ex1}
\begin{figure}[ht]
    \begin{center}
    \centerline{\includegraphics[scale=0.8]{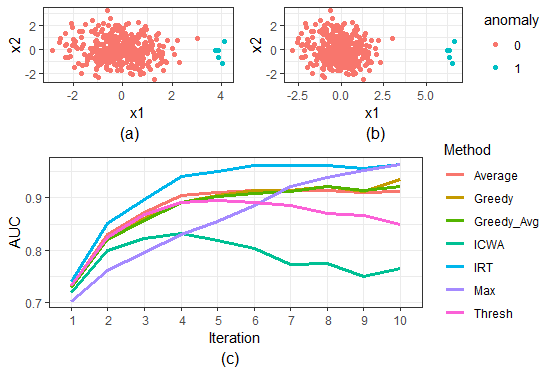}}
    \caption{Experiment 1: the anomalies moving out in each iteration. (a) The $X_1X_2$ plane in the $5^{\text{th}}$ iteration. (b) The $X_1X_2$ plane in the $10^{\text{th}}$ iteration. (c) Comparison of the AUC scores using IRT and 6 benchmark ensemble methods. }
    \label{fig:EX1}
    \end{center}
\end{figure}

The first experiment consists of two normal distributions in $\mathbb{R}^6$ with one distribution signifying the anomalies, slowly moving out of the main distribution with each iteration. We consider 5 anomalies and 400 non-anomalies. In each dimension the non-anomalies are distributed as $\mathcal{N}\left(0, 1\right)$. In the first dimension, the anomalies are distributed as $\mathcal{N}\left( 2 + (i-1)/2, 0.2\right)$ where $i$ denotes the iteration of the experiment.  Thus, in the first iteration the anomalies start with $\mathcal{N}\left( 2, 0.2\right)$ and in the $10^{\text{th}}$ iteration they are distributed as $\mathcal{N}\left( 6.5, 0.2\right)$ in the first dimension.   Figures~\ref{fig:EX1}(a) and (b) show  the anomalies in the first two dimensions in the $5^{\text{th}}$ and the $10^{\text{th}}$ iterations.  In other dimensions the anomalies are distributed similar to the non-anomalies.  A lower standard deviation is chosen for the anomalous distribution in dimension 1,  so that they are apart from the non-anomalies for higher iterations. In order to account for randomness, we repeat each iteration 10 times. 

Figure~\ref{fig:EX1}(c) shows the average AUC over 10 repetitions for each iteration. We see that the IRT ensemble achieves better results compared to the other 6 ensemble methods. Student's t-tests conducted on the differences in performance values, such as IRT performance - Greedy performance, are given in Table~\ref{tab:datExperiments}. All t-tests were statistically significant with $p$-values less than $10^{-6}$ in favor of the IRT ensemble. From Table~\ref{tab:datExperiments}  we see that IRT ensemble achieves better performance compared to the benchmarks.

\begin{table}[!t]
	\centering
	\caption{Student's t-test results of the performance differences (\%) for the synthetic experiments}
	\footnotesize
	\begin{tabular}{lrrr}
		\toprule
    Performance Difference	&  EX1 & EX2 & EX3	\\	
        \midrule
        IRT - Average &  3.73  & 4.63 & 4.86\\
        IRT - Greedy  &  3.84 & 4.63 & 6.95\\
        IRT - Greedy\_Avg & 4.97 &  4.97 & 5.95\\
        IRT - ICWA &  13.26 & 9.99  & 16.98\\
        IRT - Max & 5.78 & 8.58 & 10.20\\
        IRT - Thresh & 6.10 & 6.83 & 7.52  \\
 	 \bottomrule
	\end{tabular}
	\label{tab:datExperiments}
\end{table}

\subsection{Experiment 2}\label{sec:ex2}
\begin{figure}[ht]
    \begin{center}
    \centerline{\includegraphics[scale=0.9]{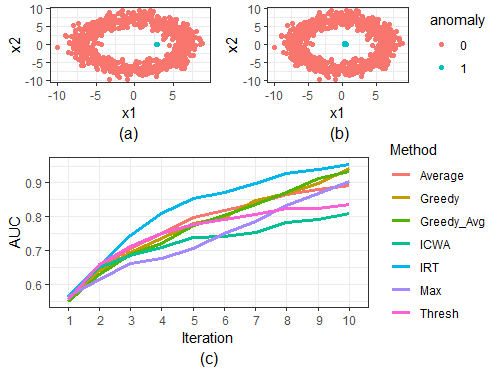}}
    \caption{Experiment 2: the anomalies moving into the annulus in each iteration. (a) The $X_1X_2$ plane in the $5^{\text{th}}$ iteration. (b) The $X_1X_2$ plane in the $10^{\text{th}}$ iteration. (c) Comparison of the AUC scores using IRT and 6 benchmark ensemble methods. }
    \label{fig:EX2}
    \end{center}
\end{figure}

For this experiment we consider data points in $\mathbb{R}^4$ that are distributed  in an annulus in the first two dimensions and normally distributed with mean 0 and standard deviation 1 in the third and fourth dimensions. Similar to the previous experiment we consider two distributions with the anomalous distribution moving into the center. The anomalies only differ in the first two coordinates $x_1$ and $x_2$ with $x_1 \sim \mathcal{N}\left( 5 - (i-1)/2, 0.1 \right)$ and $x_2 \sim \mathcal{N}(0, 0.1)$. We consider 805 points, which includes 5 anomalies. Figures~\ref{fig:EX2}(a) and (b) show the data on the $X_1X_2$ plane in its $5^{\text{th}}$ and $10^{\text{th}}$ iterations. 

Figure~\ref{fig:EX2}(c) shows the AUC results for each iteration of Experiment 2. We see that the IRT ensemble outperforms the other six ensembles. This is further confirmed by the Student's t-tests conducted on the AUC results. All t-tests were statistically significant with $p$-values less than $10^{-10}$. From Table~\ref{tab:datExperiments} we see that the IRT ensemble performs better than the other ensembles by 4.6\% to 9.9\%.  



\subsection{Experiment 3}\label{sec:ex3}
\begin{figure}[ht]
    \begin{center}
    \centerline{\includegraphics[scale=0.9]{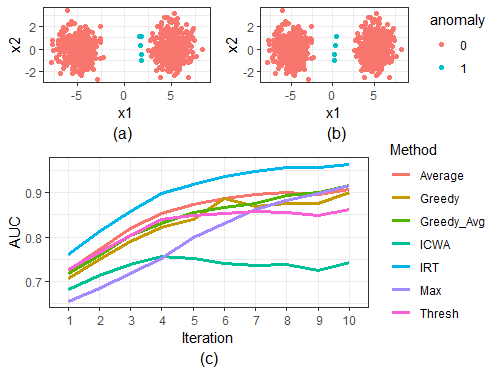}}
    \caption{Experiment 3: the anomalies moving into the trough in each iteration. (a) The $X_1X_2$ plane in the $5^{\text{th}}$ iteration. (b) The $X_1X_2$ plane in the $10^{\text{th}}$ iteration. (c) Comparison of the AUC scores using IRT and 6 benchmark ensemble methods. }
    \label{fig:EX3}
    \end{center}
\end{figure}

For this experiment we consider two normal distributions in $\mathbb{R}^6$, each having $400$ points centered at $(-5, 0, 0, 0, 0, 0, 0, 0)$ and $(5, 0, 0, 0, 0, 0, 0, 0)$ with standard deviation 1 in each dimension. This is a bimodal distribution in the $x_1$ coordinate.  A set of 5 anomalies, distributed normally with  mean $ 3 - (i-1)*0.3$  and standard deviation $0.2$ in the $x_1$ coordinate, where $i$ denotes the iteration, moves into the trough with each iteration. In other dimensions all data points are distributed as $\mathcal{N}\left(0, 1\right)$.  Figures~\ref{fig:EX3}(a) and (b) show the data points on the $X_1X_2$ plane in its $5^{\text{th}}$ and $10^{\text{th}}$ iterations. As in previous experiments, we repeat each iteration 10 times to account for randomness. 

Figure~\ref{fig:EX3}(c) shows the AUC results of all 10 iterations.   From the first iteration the IRT ensemble surpasses the other ensemble methods. Table~\ref{tab:datExperiments} gives the t-test results of the performance differences, which are all statistically significant with $p$-values less than $10^{-14}$. The IRT ensemble performs better by 4.6\% to 16\%. 



Table~\ref{tab:IRTParas} gives the discrimination, difficulty and scaling parameters of the IRT model for the example in Section~\ref{sec:example} and the last iteration of the 3 experiments EX1 - EX3. For the experiments, we see that LOF, LDF and KNN-AGG have higher discrimination parameters showing that they contribute more to the latent trait computation and thus the anomaly ensemble score. In contrast, the output of LDOF, KDEOS and INFLO are discounted. 

\section{Real-world experiments}\label{sec:real}
We now evaluate our IRT ensemble on 12433 publicly available anomaly detection datasets \citep{datasets}, which have been prepared from 119 source datasets by downsampling different classes. The minority class in each downsampled dataset is considered anomalous. Similar to the synthetic experiments, we compute the anomaly scores using the methods KNN-AGG, LOF, COF, INFLO, KDEOS, LDF and LDOF and evaluate the performance using AUC. We conduct this experiment twice with different parameter settings for the AD methods. The first parameter setting uses small values of $k$ and the second uses large values of $k$. We refer to these two takes of the experiment as T1  and T2: 
\begin{enumerate} 
\item \textbf{T1} uses the default  values for $k = k_{\min} = 5$ and $k_{\max} = 10$, and 
\item \textbf{T2} uses $k = k_{\min} = \max(\lceil N/10\rceil , 50 )$ and $k_{\max} = k +10$, making $k$ larger for the second take of the experiment. The quantity $N$ denotes the number of observations in the dataset.  \end{enumerate}

We used a high performance computing facility to run T1 and T2. For each dataset we allowed a maximum time limit of 60 minutes to complete the task. While T1 was completed by all datasets, T2 was completed only by 11194 datasets belonging to 107 dataset sources due to increased computations required by  large $k$ values.  

\subsection{By individual dataset}\label{sec:bydataset}

\begin{figure}[!ht]
    \begin{center}
    \centerline{\includegraphics[scale=0.8]{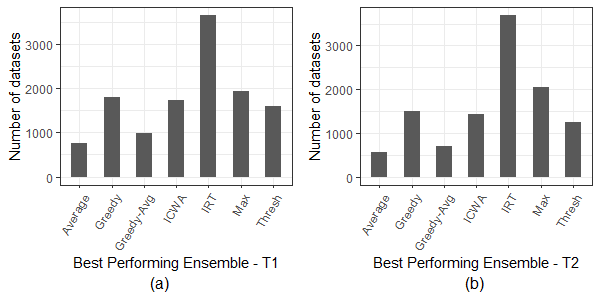}}
    \caption{The number of datasets grouped by the best performing ensemble for T1 and T2. }
    \label{fig:MaxPerf0}
    \end{center}
\end{figure}

Each dataset gives its best performance for a certain ensemble.  Figure~\ref{fig:MaxPerf0} shows the number of datasets grouped by the best performing ensemble for T1 and T2. Table~\ref{tab:bestalgobydataset} gives the percentages of datasets for the best performing ensemble for these two tasks. We see that for both T1 and T2, IRT ensemble surpasses the other ensembles by a big margin. 

\begin{table}[!ht]
	\centering
	\caption{Percentage of datasets for the best performing ensemble for T1 and T2}
	\footnotesize
	\begin{tabular}{lrrrrrrr}
		\toprule
    Experiment	 & Average & Greedy & Greedy-Avg & ICWA &  IRT & Max & Thresh	\\	
        \midrule
        T1  &  6.0 & 14.4  &    8.0   &    13.9 &  \textbf{29.4}  & 15.5   &    12.7\\
        T2  &  5.1 & 13.4  &    6.3  &   12.7 &  \textbf{33.0}   &  18.3   &    11.1 \\
 	 \bottomrule
	\end{tabular}
	\label{tab:bestalgobydataset}
\end{table}

\subsection{By dataset source}\label{sec:bydatasetsource}
\begin{figure}[!ht]
    \begin{center}
    \centerline{\includegraphics[scale=0.8]{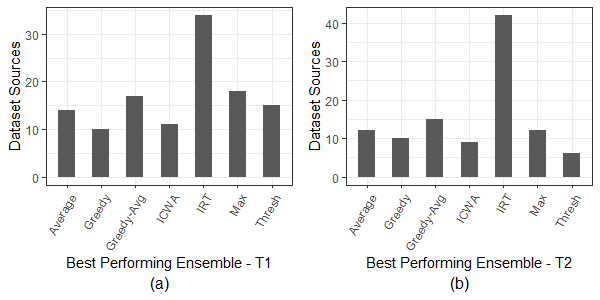}}
    \caption{Best performance for dataset sources by the ensemble method. }
    \label{fig:MaxPerf1}
    \end{center}
\end{figure}

Each dataset is obtained from a source dataset using a downsampling process. As each dataset source has many variants, we compute the mean AUC by dataset source.  For each dataset source, the ensemble with the highest mean AUC is called the best ensemble for that dataset source. Table~\ref{tab:bestalgobydatasetsource} gives the best ensemble performance percentages by dataset source and Figure~\ref{fig:MaxPerf1} gives the number of dataset sources grouped by the  best performing ensemble. Again, we see that IRT outperforms other algorithms. From Tables~\ref{tab:bestalgobydataset} and~\ref{tab:bestalgobydatasetsource} we see that IRT, Max and Thresh algorithms generally perform similarly both at the dataset and the dataset source level. We also note that Average and Greedy-Avg algorithms have larger percentages at the dataset source level compared to the dataset level, while Greedy and ICWA  have larger percentages at the dataset level. 


\begin{table}[!ht]
	\centering
	\caption{Percentage of dataset sources for the  best performing ensembles for T1 and T2}
	\footnotesize
	\begin{tabular}{lrrrrrrr}
		\toprule
    Experiment	 & Average & Greedy & Greedy-Avg & ICWA &  IRT & Max & Thresh	\\	
        \midrule
        T1  &  11.8 & 8.4 & 14.3 & 9.2 & \textbf{28.6} & 15.1 & 12.6 \\
        T2  &  11.3 & 9.4 & 14.2 & 8.5 & \textbf{39.6} & 11.3 & 5.7\\
 	 \bottomrule
	\end{tabular}
	\label{tab:bestalgobydatasetsource}
\end{table}

\subsection{Statistically significant top 2 ensembles by dataset source }\label{sec:statssigdatasetsource}
Next, we perform significance testing to further analyze these results. For each dataset source, we select the top 2 algorithms and conduct a Student's t-test to check if there is a significant difference between them. That is, if the mean performance of the best algorithm is $\mu_1$ and the mean performance of the second best algorithm is $\mu_2$, we conduct a hypothesis test with the null hypothesis $H_0: \, \mu_1 = \mu_2$ against the alternative $H_a: \, \mu_1 > \mu_2$, for each dataset source. We use a significance level $\alpha  = 0.05$. 

\begin{figure}[!ht]
    \begin{center}
    \centerline{\includegraphics[scale=0.8]{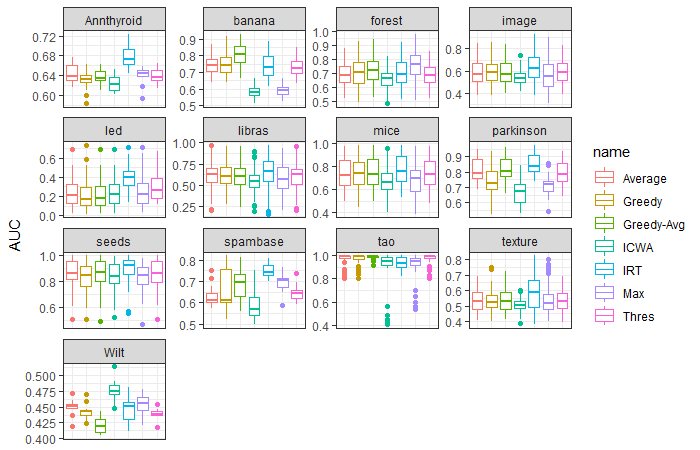}}
    \caption{T1 significant dataset sources. The IRT ensemble performs significantly better for annthyroid, image, led, libras, mice, parkinson, seeds, spambase and texture dataset sources. Greedy-Avg performs better for banana and tao.  Max performs better for forest and ICWA performs better for wilt2.  }
    \label{fig:T1significant}
    \end{center}
\end{figure}

To give some perspective of this statistical analysis, we conduct a simulated experiment. We simulate equal performance values for 7 methods for 1190 dataset sources, which is 10 times the number of dataset sources we have. For each dataset source we consider 100 datasets. Then, for each dataset source we compare the top 2 performing methods with a significance level $\alpha = 0.05$. For all 1190 dataset sources we obtained 8  sources with significant differences between the top 2 performing methods. These are false positives, since we simulated equal performance values for all methods. When we repeat this simulation 30 times, we get a mean false positive rate of 8.7 with a standard deviation of 2.89. Therefore, for 119 dataset sources we can expect $0.87 \pm 0.29$ false positives. 


We obtain 13 dataset sources with significant results in favor of the best algorithm for T1. For T2, 10 dataset sources had significant results in favor of the best algorithm.  Table~\ref{tab:sigsources} gives these results. For T1, the IRT ensemble significantly outperforms others in 9 of the 13 dataset sources. Greedy-Avg performs best for 2 dataset sources and Max and ICWA perform best for 1 dataset source each. For T2, IRT gives the best results for 7 out of 10 dataset sources while Max performs best for 2 dataset sources and Greedy performs best for 1 dataset source.  We note that the number of competing algorithms directly affect the number of dataset sources with significant performance differences between the top 2 methods. For example, when we compare IRT with Max, which is the next best algorithm (Figure~\ref{fig:MaxPerf1}(a)) for T1 in terms of dataset sources, we obtain 39 dataset sources with significant differences, of which 32 are in favor of IRT. 



\begin{table}[!ht]
	\centering
	\caption{T1 and T2 significant dataset sources and the best algorithms}
    \scriptsize
    \begin{tabular}{P{2cm}P{2cm}r|P{2cm}P{2cm}r}
		\toprule
		\multicolumn{3}{c|}{T1} &  \multicolumn{3}{c}{T2} \\
       Source &	Best Algorithm & p-value &  Source &	Best Algorithm & p-value\\
        \midrule
        annthyroid   &     IRT & $ < 10^{-5}$  & cardio10 & Greedy & 0.031 \\
           banana & Greedy-Avg  & $ < 10^{-4}$ & cardiotocography &  IRT & 0.042 \\    
           forest &       Max   &   0.023 & ecoli & IRT & 0.001 \\
            image &       IRT   & $ <10^{-4}$  & first & IRT & 0.020  \\ 
              led &       IRT   & $ <10^{-16}$ & image  & IRT  &  $ <10^{-3}$ \\  
           libras &       IRT   &  0.017   & led & Max &  < $10^{-7}$  \\   
             mice &       IRT   &  0.005 & phoneme & IRT & 0.029 \\
        parkinson &       IRT   &  0.038 & robot5 & IRT &  $ <10^{-3}$ \\
            seeds &       IRT   & 0.025 & spambase & Max & 0.034 \\
         spambase &       IRT   &  $< 10^{-4}$   & texture & IRT &  0.003 \\
              tao & Greedy-Avg  &  0.013 & & &  \\
          texture &       IRT   & $< 10^{-6}$   & & &  \\
            wilt2  &     ICWA    &   0.003   & & &  \\
    \end{tabular}
    \label{tab:sigsources}
\end{table}

\begin{figure}[!ht]
    \begin{center}
    \centerline{\includegraphics[scale=0.8]{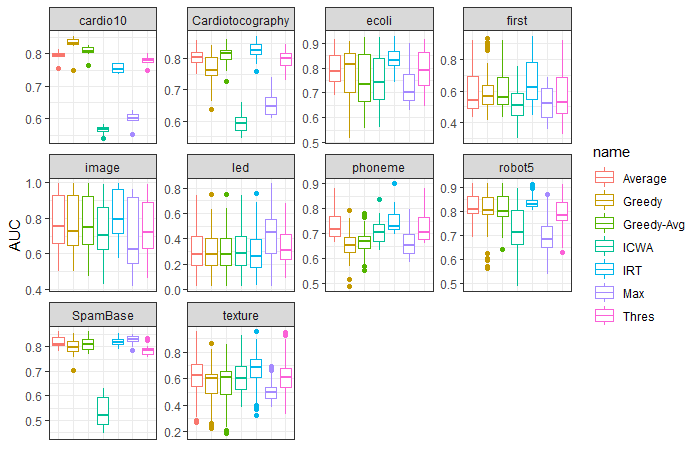}}
    \caption{T2 significant dataset sources. The IRT ensemble performs significantly better for cardiotocography, ecoli, first, image, phoneme, robot5 and texture dataset sources. Max performs better for led and spambase while Greedy performs better for cardio10.  }
    \label{fig:T2significant}
    \end{center}
\end{figure}

\begin{figure}[!ht]
    \begin{center}
    \centerline{\includegraphics[scale=0.85]{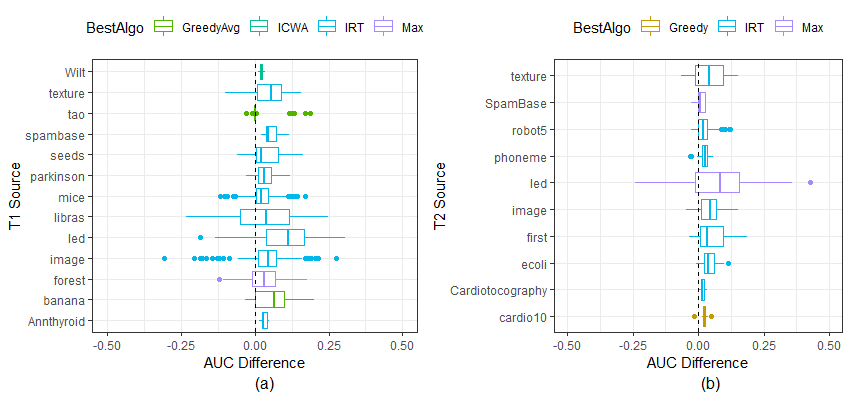}}
    \caption{AUC differences between the top 2 methods for each dataset source for T1 and T2  }
    \label{fig:AUCDiffT1andT2}
    \end{center}
\end{figure}

Figures~\ref{fig:T1significant} and~\ref{fig:T2significant} show the box plots for all 7 ensemble methods for the significant dataset sources for T1 and T2. The randomness of the downsampling process used in AD dataset preparation has resulted in high IQR and standard deviation performance values for certain dataset sources. This is because when we randomly downsample a certain class and label observations of that class as anomalies, there is a chance that these observations are not actually anomalous. Such a dataset would yield low performance values for all AD methods, making the AD ensemble methods perform poorly on that dataset. This is clearly seen in Figure~\ref{fig:T2significant}. The \textit{image} datasets have larger IQR values compared to \textit{cardiotocography} and the AUC values for  \textit{image} datasets range from 0.5 to 1 for many ensemble methods. In fact, large IQR or standard deviation values are common in AD evaluation studies as a result of the downsampling process. For example, we see large standard deviations per dataset source in  studies conducted by \cite{Campos2018} and \cite{Chiang2017}. But, because we conduct pairwise t-tests between the top 2 methods for each dataset source, the difference in AUC values by dataset is considered. As we see in Figure~\ref{fig:AUCDiffT1andT2} the differences between the top 2 methods do not exhibit large IQR values compared to previously. For example, we see that the large IQR values for \textit{image} datasets in Figure~\ref{fig:T2significant} is considerably reduced in Figure~\ref{fig:AUCDiffT1andT2}(b). Even the relatively large IQR values present in datasets \textit{libras} and \textit{led} in  Figure~\ref{fig:AUCDiffT1andT2} are smaller than their original counterparts in Figures~\ref{fig:T1significant} and~\ref{fig:T2significant}. Generally, we obtain much lower IQR values for most dataset sources by using pairwise differences between the top 2 ensemble methods. We also see that the individual boxplots are mostly 
situated in the greater than zero region, which explains why these dataset sources are significant.

\subsection{Exploring correlation}
\begin{figure}[!ht]
    \begin{center}
    \centerline{\includegraphics[scale=0.85]{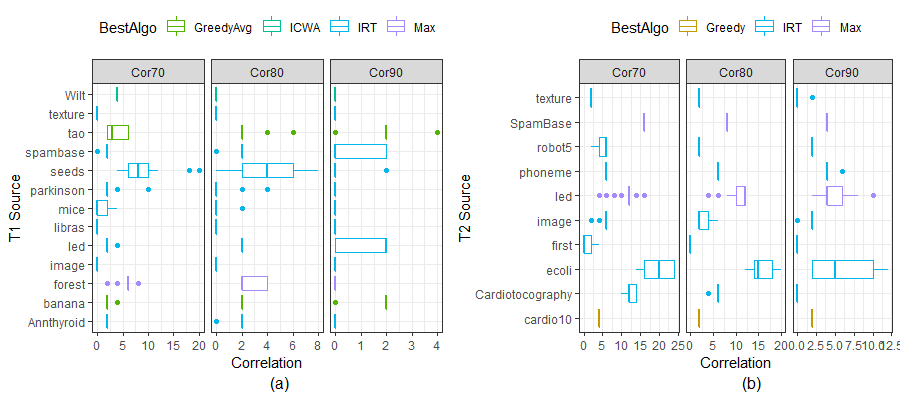}}
    \caption{Correlation counts for significant dataset sources in T1 and T2  }
    \label{fig:CorT1andT2}
    \end{center}
\end{figure}

To better understand the performance differences of AD ensemble methods, we investigate the correlation matrix $\cor(Y)$ of each dataset, where matrix $Y$ contains the original anomaly responses from the 7 AD methods. For each dataset we compute $\cor(Y)$ and count the number of entries greater than 0.9, 0.8 and 0.7 respectively. From these counts, we deduct the number of ones on the diagonal, as this is redundant information. We denote these values by \textit{Cor90}, \textit{Cor80} and \textit{Cor70}, and collectively refer to them as \textit{correlation counts}. Thus, if the anomaly responses of different AD methods are highly correlated for a given dataset, it would result in high correlation counts. Figure~\ref{fig:CorT1andT2} shows the boxplots of Cor70, Cor80 and Cor90 for significant datasets in T1 and T2. From these plots we see that IRT performs well on datasets with both low and high correlation counts. But we also see that only IRT performs well on datasets with very low correlation counts. For example in Figure~\ref{fig:CorT1andT2}(a) the dataset sources with zero median values for Cor70, Cor80 and Cor90 are \textit{texture}, \textit{mice}, \textit{libras} and \textit{image}. And, IRT is the best algorithm for these dataset sources. Similarly, in Figure~\ref{fig:CorT1andT2}(b) the dataset sources with zero median values for Cor90 are \textit{texture}, \textit{first} and \textit{cardiotocography}. Again, IRT is the best ensemble method for these dataset sources. 

\begin{figure}[!ht]
    \begin{center}
    \centerline{\includegraphics[scale=0.8]{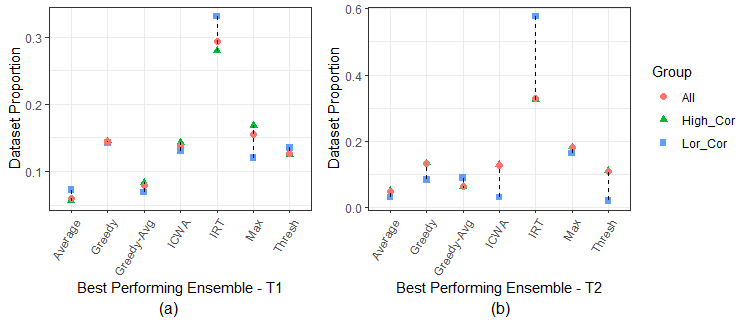}}
    \caption{High\_Cor, Low\_Cor and overall dataset performance for different ensemble algorithms for T1 and T2. }
    \label{fig:CorT1andT22}
    \end{center}
\end{figure}

These observations may indicate a possible strength of the IRT ensemble for datasets with low correlation counts. However, the significant dataset sources are heavily dominated by IRT in terms of best performance, which can induce a random result in favor of IRT for datasets with low correlation counts. Therefore, we explore this further using the full data repository of 12000+ datasets. For both T1 and T2, we focus on the datasets with $Cor70 = 0$, that is, the correlation matrix of the anomaly responses $\cor(Y)$ of that dataset has no correlation values above 0.7, which also implies that $Cor80 = Cor90 = 0$. Let us call this group of datasets \textit{Low\_Cor} and the rest \textit{High\_Cor}. We compare the proportion of the best performing algorithms for \textit{Low\_Cor} and \textit{High\_Cor} datasets for both T1 and T2. Figure~\ref{fig:CorT1andT22} shows these results. For T1, there were 3394 \textit{Low\_Cor} datasets and for T2 there were 189. From Figures~\ref{fig:CorT1andT22}(a) and (b) we see that IRT not only has the highest, best performance proportion when we consider all datasets, it has even better performance for \textit{Low\_Cor} datasets.

Going back to traditional IRT, correlation values can be low in educational testing. For example, the correlation of marks between a very easy question and a very difficult question can be extremely low. By computing difficulty and discrimination of test items, IRT has built in capability to accommodate low correlated items. We see this as a particular strength of IRT when computing an ensemble score.

\section{Conclusions}\label{sec:conclusions}
Most ensemble techniques use the ground truth or the class labels to construct an ensemble score. As the ground truth is not known in unsupervised anomaly detection, constructing an ensemble using a set of heterogeneous AD methods is a challenge. We have introduced a novel ensemble framework for anomaly detection using Item Response Theory that uses IRT's latent trait, which denotes an underlying characteristic, to compute  the ground truth. Latent trait modeling has not been used before in AD ensembles to the best of our knowledge. 

Using 6 other AD ensemble techniques, we evaluated the IRT ensemble on synthetic and real-world data. The results demonstrated that our method is very attractive for anomaly detection ensembling tasks. Furthermore, we see that the IRT ensemble has a unique strength for datasets with low correlation values in $\cor(Y)$.  This is advantageous when constructing an ensemble using diverse methods. As future research avenues, we plan to consider IRT ensembles for other unsupervised machine learning tasks and to extend this framework for supervised tasks. 

\section*{Supplementary Material}\label{sec:supmat}
The IRT ensemble and the other AD ensembles are available in the R package \texttt{outlierensembles} \citep{outlierensembles}. The programming scripts are available at \url{https://github.com/sevvandi/supplementary_material/anomalyensembles}.  




\bibliography{citations}
\bibliographystyle{agsm}


\end{document}